\documentclass[]{rsos}

\usepackage{url}
\usepackage{pdfpages}
\usepackage{comment}
\usepackage{amsthm}
\usepackage{tabularx}
    \newcolumntype{L}{>{\raggedright\arraybackslash}X}

\newtheorem*{definition}{Definition}
    
\begin{document}

\title{The Fundamental Principles of Reproducibility}

\author{
Odd Erik Gundersen}

\address{
Department of Computer Science, Norwegian University of Science and Technology, Trondheim, Norway\\
Tr{\o}nderEnergi AS, Trondheim, Norway  }

\subject{Reproducibility, artificial intelligence}


\corres{Odd Erik Gundersen\\
\email{odderik@ntnu.no}}

\begin{abstract}
Reproducibility is a confused terminology. 
In this paper, I take a fundamental view on reproducibility rooted in the scientific method. 
The scientific method is analysed and characterised in order to develop the terminology required to define reproducibility.
Further, the literature on reproducibility and replication is surveyed, and experiments are modeled as tasks and problem solving methods.
Machine learning is used to exemplify the described approach.
Based on the analysis, reproducibility is defined and three different degrees of reproducibility as well as four types of reproducibility are specified.

\end{abstract}


\begin{fmtext}
\section{Introduction}

I propose to consider the question "What is repro-ducibility?"
As reproducibility is such a central concept in science, one would think that it would be clearly defined. 
However, this is not the case. 
Reproducibility is an elusive concept \cite{Plesser_2018}. 
It has no single commonly agreed upon definition. 
Rather, it has many different, and each of these captures central, but different properties of it. 
In that regard, the situation is similar to the one in the parable of \emph{the Blind Men and the Elephant}.
Each blind man conceptualizes reproducibility by touching a part of it, and each of their descriptions are different as none of them describe the concept as a whole.


The lack of a definition of reproducibility is an issue when considering automating the scientific endeavour. 
This has been coined the next grand challenge for artificial intelligence (AI).
Yolanda Gil \cite{Gil_2020} challenged the AI community in her Presidential address at AAAI 2020 to develop an AI that could be a co-author of a paper, capable of formulating research questions and generate novel contributions by 2040.
\end{fmtext}
\maketitle 
Hiroaki Kitano \cite{Kitano_2016} goes one step further by proposing AI's that coordinate collaborating intelligent agents -- both humans and computers -- that endeavour in scientific discovery, and as such it could be a potential co-recipient of one of the Nobel prizes in science.
A subtask then is to formulate the scientific method in such a way that computers can execute it. 
This means representing the tasks the scientific method is comprised of at the right abstraction level and implementing methods to solve them. 
In order to fully automate scientific discovery, we must make reproducibility tangible as well. 

In order to analyse and understand reproducibility, it is viewed it in the light of the scientific method. 
The scientific method is modeled and machine learning is used as an example to illustrate this model. 
Based on this, reproducibility is defined. 
The definition is general and applies to all sciences, despite relying on examples from machine learning.
One might argue that I am just another blind man, as I analyze the concept of reproducibility through the lens of one research field.
On the contrary, I will argue that computer science is no less than exactly the right lens to see and try to understand reproducibility through. 
An important aspect of computer science is to represent the world and develop step-by-step processes of how to solve problems on a computer.
Given that computers let you to make more mistakes faster than any invention in human history, one has to be very precise and careful when describing the task it is to perform. 
Because of this, computer science has developed many tools for analysing and understanding the problem that is to be solved before solving it. 
Examples include UML \cite{booch2005} for visualising the design of computer systems, Petri nets \cite{rozenberg1996} for the description of distributed systems and Behaviour trees \cite{colledanchise2018} for modeling plan execution. 
Task modeling of problem solving \cite{Chandrasekaran_1990,Steels_1990} is of particular interests, as the scientific method can be described as a task and decomposed further into subtasks using problem solving methods. 
This work is build upon this when analysing reproducibility. 




\section{The Scientific Method and Machine Learning}
The discussion and analysis of reproducibility requires a shared view on the scientific method. 
So, this section outlines a process view of it and provides an example from machine learning.
The scientific method is a systematic method of acquiring knowledge about the world through observations in order to ensure that our beliefs about the world are as certain as they can be.
This is done through the testing of hypotheses that typically are proposed on the basis of a scientific theory.
A hypothesis is a statement about the world that is either true or false, and it is tested by conducting an experiment.
Reproducibility is about confirming the results of a past experiment by conducting a \emph{reproducibility experiment}.
Theories provides explanations of the natural world, and scientific theories are in addition falsifiable \cite{Popper_1963}. 
The scientific method is an empirical method, and Cohen \cite{Cohen_1995} distinguishes between four different classes.
Exploratory and assessment studies are conducted to identify and suggest possible hypotheses and are as such performed to provide input to the scientific method. 
Manipulation and observational studies test explicit and precise hypotheses.


\begin{figure}[!t]
\centering\includegraphics[width=\textwidth]{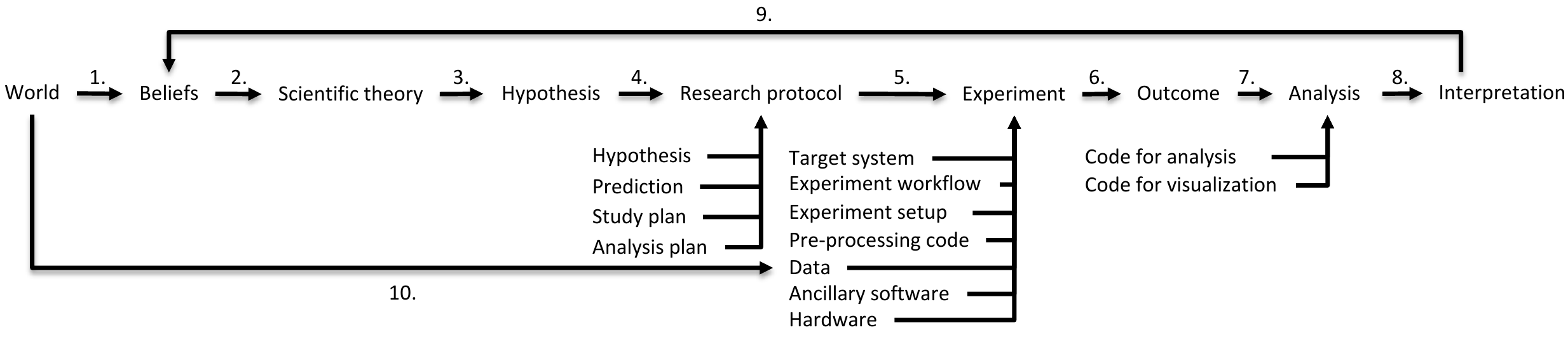}
\caption{The scientific method as a ten step process: 1) \emph{observe the world to form beliefs about it}; 2) \emph{explain causes and effects by forming a scientific theory}; 3) \emph{formulate a genuine test of the theory}; 4) \emph{design an experiment to test the theory}; 5) \emph{implement the experiment}; 6) \emph{conduct the experiment}; 7) \emph{analyse the outcome}; 8) \emph{interpret the analysis}; 9) \emph{update beliefs according to the result}; and 10) observe the world systematically.}
\label{fig:sientific_method-cs}
\end{figure}

Figure~\ref{fig:sientific_method-cs} illustrates a process view of the scientific method.
An experiment documenting a deep learning neural network for image classification \cite{Ciregan_2012} is used as the example.
The arrows in the figure indicate the sub-processes that the scientific method is comprised of. 
Each of the sub-processes are referred to with a number, and each sub-process is given a brief description below. 
The nouns that the arrows point at describe the outputs of the sub-processes. 
These are the research artifacts. 
The ten sub-processes are:
\begin{enumerate}
    \item \emph{Observe the world to form beliefs about it:} 
    We typically observe the world in an unsystematic and unstructured way, so that the observations have a sampling bias. 
    Applying the scientific method helps reduce this bias.
    Exploratory and assessment studies are conducted to form these beliefs.
    \item \emph{Explain causes and effects by forming a scientific theory:} 
    The scientific theory underlying the example is that deep neural networks are models of the brain, although simple ones, and as such intelligence could emerge from them \cite{Minsky_1988,Downing_2015}.
    \item \emph{Formulate a genuine test of the scientific theory as a hypothesis:}
    The hypothesis that is tested in the example is that the performance of biological inspired deep convolutional neural networks is competitive with human performance on computer vision benchmark tasks.
    \item \emph{Design an experiment to test the hypothesis and document the experiment in a research protocol:}
    Ideally, research protocols documenting the experiments should be written ahead of them being conducted \cite{chambers2013}.
    Experiment design has traditionally not been a very structured process in computer science.
    Often, making the research protocol and the experiment is an iterative process in which the experiment design, implementation and execution are done interchangeably.
    This could be considered HARKing or hypothesizing after results are known \cite{kerr1998}, which is not good practice.
    \item \emph{Implement the experiment so that it is ready to be conducted:}
    The target of research in a machine learning experiment is often a system or an algorithm such as the biologically inspired deep neural network architecture of the example. 
    In figure~\ref{fig:sientific_method-cs} this is called the target system, and it is a piece of code that is written by the researchers themselves that has some potential properties that they want to test.  
    The experiment has many more components, many of them are also code written by the researchers. 
    Often, the collected data requires pre-processing.
    In the example, the raw images are translated, scaled, rotated and distorted.
    Hyperparameters, random seeds, the learning rate, the number of epochs and so on are configured in the experiment setup. 
    Another piece of code could define the experiment workflow.
    Figure 1a in the example \cite{Ciregan_2012} illustrates the pre-processing workflow.
    Also, data must be gathered. 
    In this case the data is compiled from six different computer vision benchmarking data-sets. 
    Furthermore, the experiment must be run on some hardware, which in the example is a graphical processing units (GPU).
    Most experiments relies on ancillary software to run the experiment . 
    Ancillary software includes but are not limited to the operating system and software libraries that simplifies the execution of the experiment. 
    \item \emph{Conduct the experiment to produce outcomes:}
    Conducting a machine learning experiment typically requires executing software on a computer without any input from the outside world except for the training and test data.
    The outcomes that are produced in the example are class labels for the images in the test data from the benchmarking data-sets. 
    \item \emph{Analyse the outcomes to make an analysis:}
    The analysis typically consists of visualisations of the outcomes and metrics that are computed based on the outcomes. 
    In the example, the analysis is to compute error rates and display them in tables.
    \item \emph{Interpret the analysis:}
    The analysis has to be interpreted.
    This interpretation leads to a conclusion, and this conclusion is the result of the experiment.
    In the example, the analysis of the outcomes show that computers have lower errors than humans on these tasks, which when interpreted leads to the conclusion that deep convolutional neural networks are competitive with humans on widely used computer vision benchmark tasks.
    
    \item \emph{Update beliefs according to the interpretation:}
    Scientists update their beliefs based on trusted results and interpretations even if they are the exact opposite of previous beliefs.
    Surprising and counter-intuitive results might not be trusted immediately, but they could spur new and different experiments to increase the trust. 
    Although the analysis (low errors by deep learning methods on visual benchmarking tasks) have been reproduced many times \cite{zhao2019}, the claim that deep learning achieves super-human performance is still debated \cite{geirhos2018}.
    \item \emph{Observe the world systematically:}
    To be a trusted source of knowledge, experiments must be designed to remove biases.
    As many data-sets are biased \cite{torralba2011}, the bias can be reduced by conducting the experiment on several data-sets, similar to what is done in the example.
\end{enumerate}
It is important to note that one can test a hypothesis without basing it on predictions made from a scientific theory. 
For example, one could test the hypothesis "all swans are white". 
No theory is needed here. 
So, the second sub-process \emph{explain causes and effects} might be bypassed.

\section{A Survey of Definitions}
In this section, definitions of repeatability, replicability and reproducibility are reviewed. 
Table~\ref{tab:survey} presents a short overview of the papers that are surveyed. 
Different terms are used in these papers to describe the definitions. 
In order to compare the definitions a unified set of terms based on  Figure~\ref{fig:sientific_method-cs} is used.
A brief description of these follows: 
\begin{description}
\item \emph{Investigators:} 
The investigators are those who conduct the experiment.
Some definitions rely on whether the same or independent investigators perform the reproducibility experiment.
\item \emph{Experimental procedure:} 
The experimental procedure is the sequence of operations that has to be done in order to conduct the experiment. 
It is typically outlined in a research protocol.
\item \emph{Implementation:} 
This is the exact steps the investigators take to carry out the experimental procedures.
In an ML experiment, the experimental procedure is encoded programmatically as source code.
The source code encodes the behaviour of the target system, the experimental setup and the experimental workflow. 
\item \emph{Ancillary software:}
Ancillary software is software that is used as part of the experiment, but is not implemented by the investigators. 
It includes but is not limited to the operating system, software libraries, software frameworks and software packages such as Matlab, and it captures one dimension of what constitutes a laboratory.
\item \emph{Hardware:}
Hardware is the tools and computers required to conduct the experiment and is the other dimension that constitutes a laboratory. 
\item \emph{Data:} 
The systematic observation of the real world that are recorded as part of the experiment and stored as data.
If any pre-processing of the data is done, it has to be pre-processed in the exact same way if the observations should be considered the same.
\item \emph{Outcome:} 
In many experiments, the difference between the outcome and the observations is a subtle one as the outcome of some experiments is data. 
In such experiments, the observation is the recognition of an event in the real world and the outcome is the record of it.
In a machine learning experiment on the other hand, the input is some data that record the observations and is used by the target system to produce the outcome. 
The observations (data) are the images of hand-written digits that is input to the target system and the outcome is the labels produced by the target system.
\item \emph{Analysis:} 
The analysis is done on the outcome, which could be to compute the error or other statistics that are listed in tables or visualized as charts or graphs.
\item \emph{Interpretation:} 
The analysis is interpreted so that a conclusion can be reached. 
The conclusion constitutes the result of the experiment, and if it is conclusive, beliefs can be updated accordingly. 
\end{description}

\begin{table}[h!]
    \centering
    \begin{tabularx}{\linewidth}{|p{4cm}|L|} 
        \hline
        \textbf{Reference} & \textbf{Definition}\\ 
        \hline
            Claerbout and Karrenbach \cite{claerbout1992} and Schwab \cite{Schwab_2000} & Reproducibility is the ability to recalculate a figure from data, parameters and programs.\\
        \hline
            Buckheit and Donoho \cite{buckheitdonoho95} & Reproducibility is the ability to run software that produced figures.\\ 
        \hline
            Peng et al. \cite{Peng_2006} & Replication is confirming scientific evidence by multiple independent investigators using independent data, analytical methods, laboratories, and instruments and reproducibility is independent investigators subjecting the original data to their own analyses and interpretations.\\ 
        \hline
            Drummond\cite{drummond2009} & "Reproducibility requires changes; replicability avoids them."\\ 
        \hline
            Schmidt \cite{Schmidt_2009} & Distinguishes between two fundamental levels of replication: 1) the narrow bounded notion that is a repetition of experimental procedures, and 2) a wider notion of replication that is a test of a hypothesis or a result of earlier research with different methods. \\ 
        \hline
            Miller \cite{Miller_2010} and Currie and Svehla \cite{Currie_1994} & Repeatability is  when the same team repeats the same experiment in the same lab and reproducibility is the when another team performs the same experiment in a different lab.\\ 
        \hline
            Peng \cite{peng_2011} & Reproducibility spectrum that goes from not reproducible to gold standard depended  on whether code and data are shared and executable.\\ 
        \hline
            Stodden \cite{stodden2011trust} & Replication is the regeneration of published results from author-provided code and data while reproducibility is a more general term. \\ 
        \hline
            JGCM\footnote{Joint Committee for Guides in Metrology} \cite{JCGM_2012} & Repeatability is to keep conditions the same, while reproducibility is to change conditions of the experiment. \\ 
        \hline
            Crook et al. \cite{Crook_2013} & Distinguish between three different types of replication and reproducibility (related to simulations).\\ 
        \hline
            Gent and Kotthoff \cite{gent2013recomputation, gent2014recomputation} & Recomputation is to re-execute an computational experiment in the same environment using virtual machines containing the complete experiments.\\ 
        \hline
            Nosek and Lakens \cite{Nosek_2014} & Direct replication is the attempt to duplicate the conditions and procedures that existing theory and evidence anticipate as necessary for obtaining the effect. Reproducibility is establishing the validity of the reported findings with new data\\ 
        \hline
            Goodman et al. \cite{Goodman_2016} & Proposes new terminology, methods reproducibility, results reproducibility and inferential reproducibility, instead of using reproducibility and replication.\\ 
        \hline
            ACM\footnote{Association for Computing Machinery} \cite{ACM_2020} & Defines repeatability, replicability and reproducibility for computational experiments along the two dimensions team and experimental setup.\\ 
        \hline
            Gundersen and Kjensmo \cite{gundersen_kjensmo_2018} & Define three degrees of reproducibility based on which documentation is shared with independent researchers. \\ 
        \hline    
            NAS\footnote{National Academies of Sciences, Engineering, Medicine} \cite{national2019} & Define reproducibility to mean computational 
            reproducibility and replicability to mean obtaining consistent results across studies answering the same scientific questions. \\
        \hline
    \end{tabularx}
    \caption{A brief overview of the papers that are surveyed. 
    }
    \label{tab:survey}
\end{table}

\begin{figure}[!t]
\centering\includegraphics[width=5in]{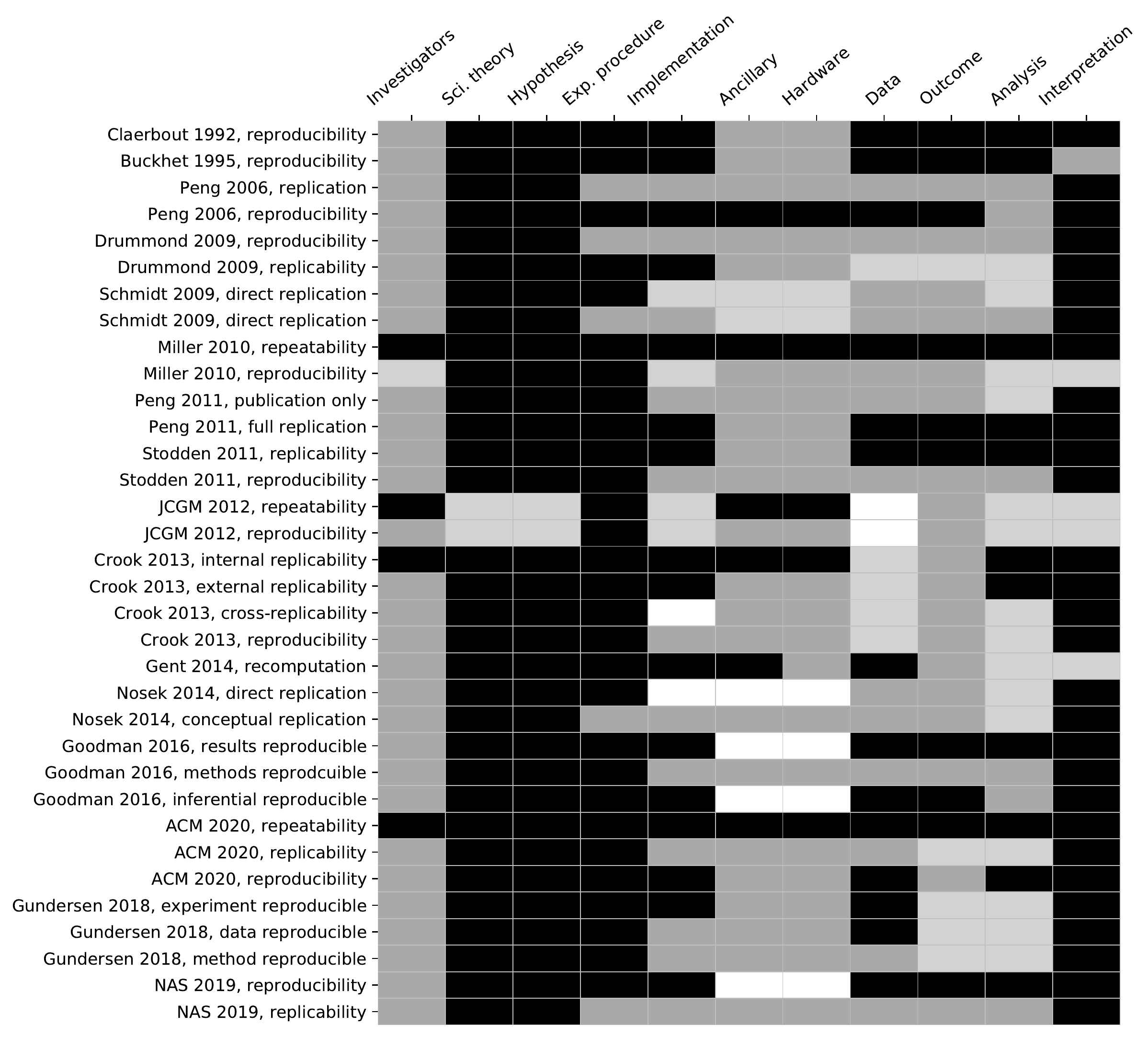}
\caption{
The color of each cell indicates what has to be kept the same (black), similar (white), different (dark grey) or not specified (light grey) between the original and the reproducibility experiments.
}
\label{fig:reproducibility_survey_heatmap}
\end{figure}

It is important to note that the terms used above are not necessarily the same ones that are used in the reviewed resources. 
This means that some subjective interpretations have been made and that these might be discussed. 
The interpretations are documented for each resource in the supplementary material, so please refer to it for details. 
Figure~\ref{fig:reproducibility_survey_heatmap} presents the definitions from the surveyed papers and their interpretation as described by the terminology above. 
The first author and publication year of the paper is listed before the term that is defined by the authors.
A term is considered more general the less variables that are kept fixed to \emph{same}.
According to the findings, the least general term is repatability. 
The authors that define repeatability as fixing the most variables to \emph{same} compared to the other definitions they make. 
From this survey, it is clear that repeatability means that the same investigators conduct the same experiment under the same conditions over again.

I will clarify two of the decisions on terminology that has been made. 
First, the definitions that require the reproducibility experiment to be conducted in same laboratory are  interpreted as requiring the same hardware and ancillary software while definitions that refer to conducting the same experiment are interpreted as executing the same experimental setup, pre-processing workflow and target system on the same data.
Second, it is not always clear what is meant by results. 
For example, what exactly is meant by re-create the results in a publication when rerunning simulation software as specified in internal replicability by Crook et al. \cite{Crook_2013}? 
Do they mean that results are re-created if the outcome produced by the simulator is be exactly the same for the reproducibility experiment?
It could also mean that the analysis of the outcome should be the same or that the same conclusion can be drawn from the interpretation of the analysis.
In this analysis, the three terms outcome, analysis and interpretation are used to distinguish between different results.
The term result of an experiment is reserved to be the conclusion that is drawn based on the interpretation of the analysis. 
Most of the surveyed definitions require that results or claims to be confirmed, which is interpreted as at least interpretations having to be the same. 

Peng et al.  \cite{Peng_2006}, Drummond \cite{drummond2009}, Stodden \cite{stodden2011trust}, Crook \cite{Crook_2013}, Peng \cite{peng_2011}, ACM \cite{ACM_2020} and NAS \cite{national2019} define both the term reproducibility and the term replication. 
Peng et al. \cite{Peng_2006}, Peng \cite{peng_2011}, ACM \cite{ACM_2020} and NAS \cite{national2019}, define replication to be the more general term.
Although close, based on this it seems that the consensus is that replication is the more general term.
When only using one term to describe different degrees of reproducibility or replication, the authors Schmidt \cite{Schmidt_2009} and Nosek and Lakens \cite{Nosek_2014} use replication while Goodman et al. \cite{Goodman_2016} and Gundersen and Kjensmo \cite{gundersen_kjensmo_2018} use reproducibility. 
Crook \cite{Crook_2013} define three different degrees of replication in addition to reproducibility, while Goodman et al. \cite{Goodman_2016} and Gundersen and Kjensmo \cite{gundersen_kjensmo_2018} propose three different degrees of reproducibility. 
They do not seem to agree much on the different levels. 

Except for the definitions of indirect replication, as provided by Schmidt \cite{Schmidt_2009} and Nosek and Lakens \cite{Nosek_2014}, and reproducibility as described by Drummond \cite{drummond2009}, all definitions of reproducibility and replication are related to conducting an experiment with the same or similar experimental procedure over again with different degrees of variation of conditions.
Indirect replication and Drummond's interpretation of reproducibility are about strengthening of the belief in hypotheses and scientific theories, which seems to collide with the notion of reproduce and copy that are the main terms used for describing reproduce and replicate. 
Hence, they seem to resemble corroboration more than reproducibility or replication.
\section{On the Difference between Corroboration and Reproducibility}

\begin{figure}[!t]
\centering\includegraphics[width=5in]{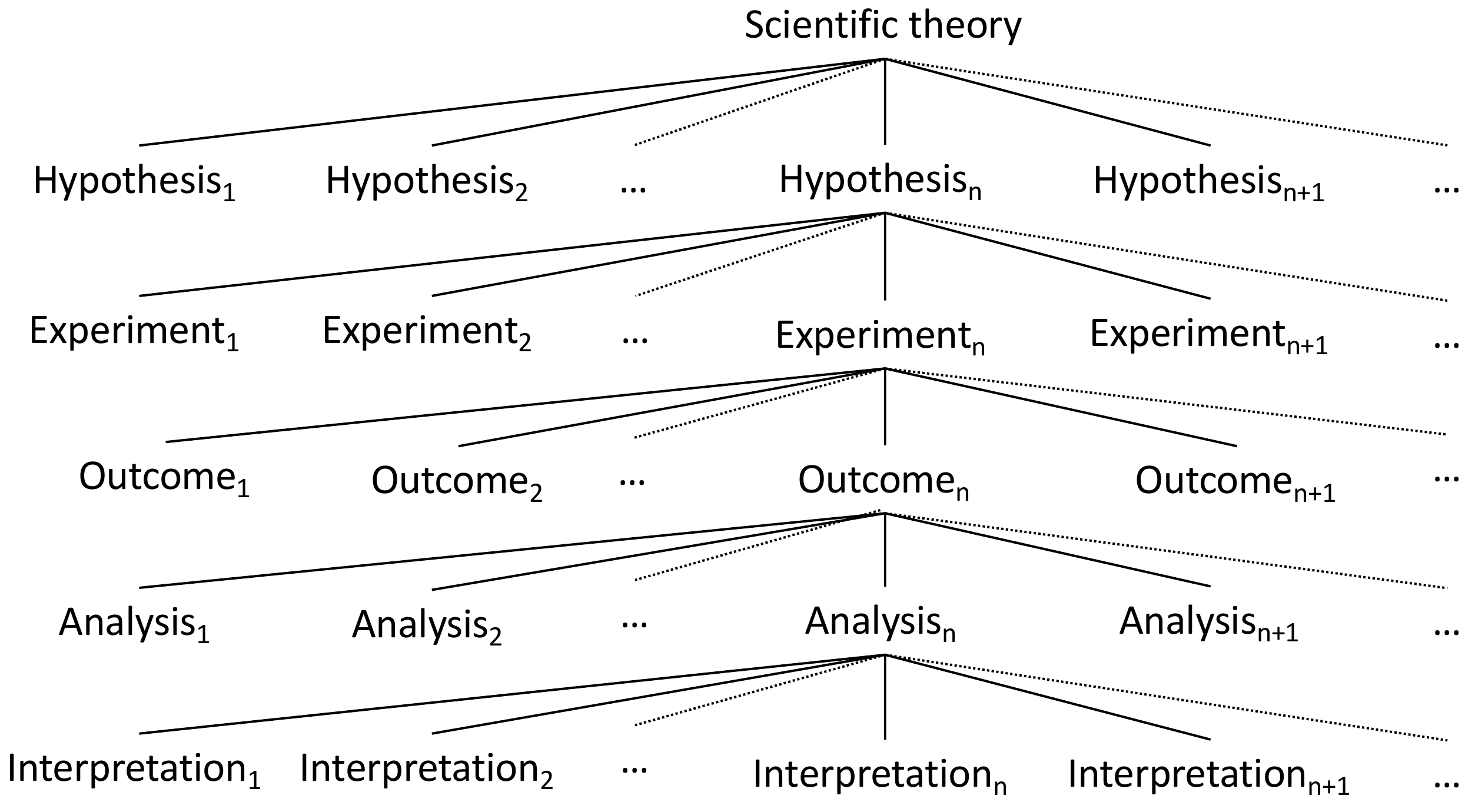}
\caption{The relationship between the concepts used for explaining the scientific method as a process.}
\label{fig:realtionsip_between_terms}
\end{figure}

According to Merriam-Webster's dictionary corroborate is \emph{to support with new evidence} or \emph{make more certain} and reproduce is to \emph{to produce again} or \emph{make a copy of}.
While reproducibility is to produce again or make a copy, corroboration is related to gathering new evidence and make more certain.
Corroboration is a more general term than reproducibility. 
Reproducibility is related to experiments, while theories and hypotheses can only be corroborated. 

A theory is a concept or a belief and hence it does not have any physical manifestations. 
However, whether it is correct or not can be tested by formulating multiple hypotheses as shown in Figure~\ref{fig:realtionsip_between_terms}.
Consider Einstein's general theory of relativity \cite{Einstein_1922}. 
The theory has been tested by formulating many different hypotheses including the deflection of light by the Sun \cite{Dyson_1920} and galaxies acting as a gravitational lenses \cite{Collett_2018}.

Each new hypothesis formulated on basis of a theory that are supported by rigorous testing adds evidence to the correctness of the theory; the theory is corroborated if supported by new hypotheses. 
Hypothetically, one could imagine that there exist a finite amount of hypotheses that could be formulated to test a theory. 
Further, we could imagine that these hypotheses are formulated and experiments that test these hypotheses are designed and conducted.
The outcomes of these experiments are analysed and interpreted. 
In the case where everyone agree that all the interpretations support the theory, we would have to consider the theory true as all evidence lead to this conclusion.
There is no more evidence that can be produced to support the theory.
Reproducing the results of any of the proposed experiments will not add any additional evidences supporting the theory. 
A theory is not reproduced, only corroborated.

Hypotheses cannot be reproduced either. 
A hypothesis is corroborated by designing new and different experiments that test the hypothesis as illustrated in Figure~\ref{fig:realtionsip_between_terms}.
Evidence is added when a new and different experiment supports the same hypothesis; the belief in the hypothesis is strengthened. 
Conducting the same experiment over again will not add new evidence to  the correctness of the hypothesis.
It will only strengthen the belief in the result of the experiment. 

Conducting an experiment several times will however make us more certain that we draw the correct conclusion from the experiment. 
When conducted several times, the same experiment can produce many different outcomes as evident by how reproducibility and repeatabililty are defined in measurement theory \cite{Miller_2010}.
However, this is not only restricted to measurements. 
Running the same experiment on a computer does not necessarily produce the same outcomes as discussed by Nagarajan \cite{Nagarajan_2019}.
Differences in outcomes could be caused by using different hardware architectures, operating systems, compiler settings \cite{Hong_2013}, intentional stochasticity in algorithms, random number generator seeds and hardware, such as conducting the experiment on a graphics processing unit (GPU) as discussed by Henderson et al. \cite{Henderson_2018}.    
A result can be analysed in many different ways by using different charts, performance metrics or error analyses and these can reveal different information that allows for different interpretation, which means that different investigators might interpret the analysis in different ways and thus reach different conclusions \cite{Goodman_2016}.


So, to conclude, our beliefs in scientific theories and hypotheses are strengthened with new evidence. 
Reproducing the results of an experiment will increase our trust in the conclusions we draw from the experiment.  
The belief in the hypothesis will not be strengthened by reproducing an experiment. 
New and different experiments are needed for that. 

\section{Experiments as Tasks}
In recent years, end-to-end machine learning has become more or less synonym with AI. 
The reason is of course the advancement of deep neural networks and the impressive results that have been achieved. 
Deep learning was the main component in the AI's that beat top level human players of Atari \cite{Mnih_2015}, go, chess, shogi \cite{Silver_2018}, poker \cite{Brown_2019} and Starcraft \cite{vinyals2019}.
These systems are called end-to-end as they learn to play these games without human intervention.
Deep models learn their own representation of the task to solve, and these representations are not easy for humans to fully understand and interpret \cite{lipton2018}. 
During the 1980's and 1990's, AI was approached in a very different way.
In expert systems, the expertise of the systems was formulated explicitly by humans. 
While this way of designing AI systems has some clear disadvantages, quite a lot of effort was put into formulate the components of expertise to develop AI systems that represented knowledge explicitly. 
One direction was to represent the knowledge as domain models, tasks and problem solving methods. 
Chandrasekaran \cite{Chandrasekaran_1990} and Steels \cite{Steels_1990} provide introductions and summaries of how to represent this expertise. 

\begin{figure}[!t]
\centering\includegraphics[width=5in]{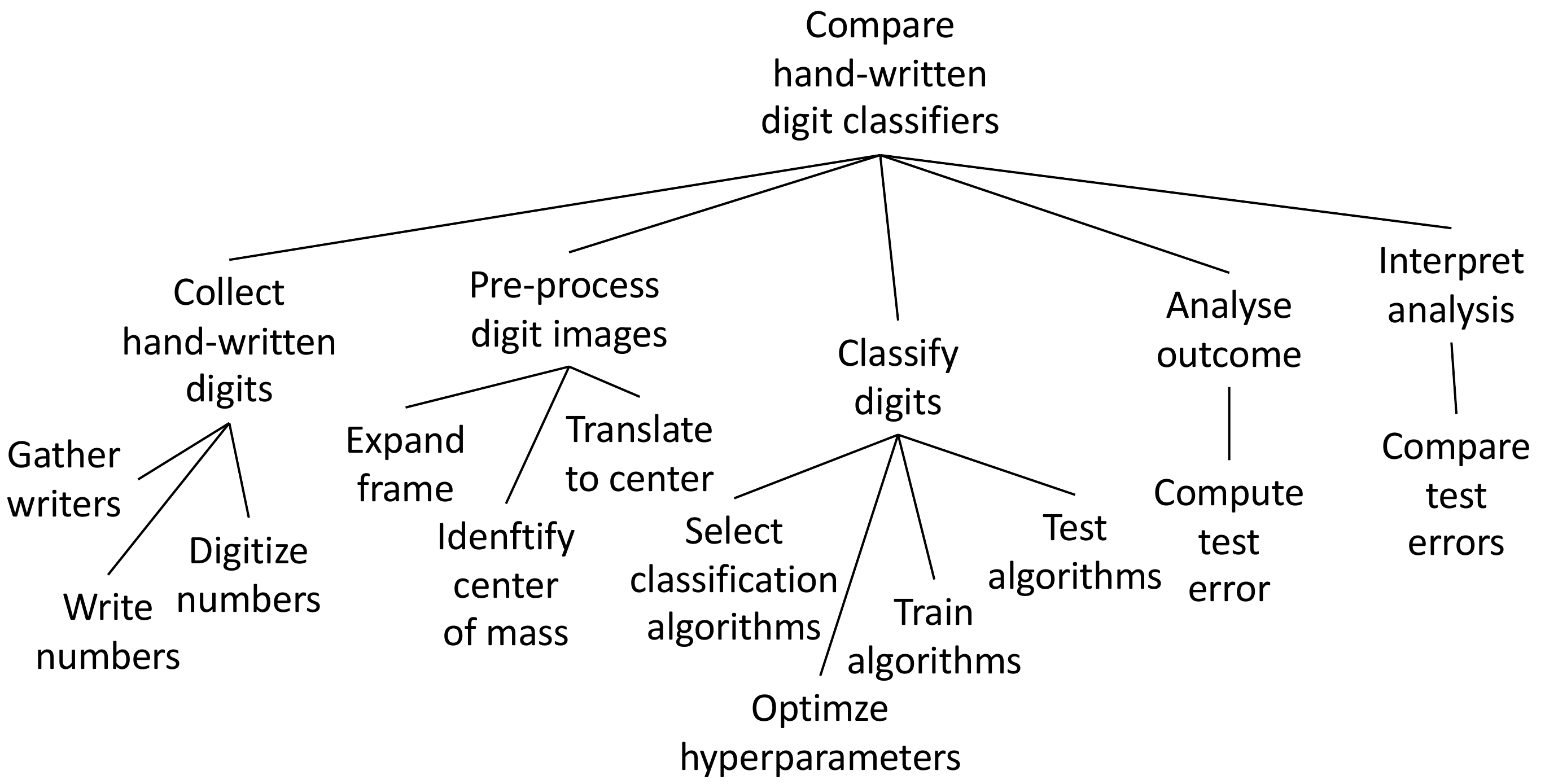}
\caption{A simplified and naive task tree for an experiment comparing hand-written digit classifiers.}
\label{fig:hand-written_digit_calssification}
\end{figure}

I propose that tasks and problem solving methods could be utilised as a tool for analysing reproducibility. 
Here, I will use the terms \emph{task} and \emph{problem solving method} as specified by Öztürk et al. \cite{Ozturk_2010}.
Tasks are characterised by their goal, input, output and a reference to the problem solving methods that complete them while problem solving methods are characterised by their input, output, the subtasks they decompose the parent task into and the control information specifies i.e. order of execution of tasks.  
An experiment could be interpreted as a task with the goal of testing a specific hypothesis. 
A task can be completed by executing a problem solving method, and there might exist several different problem solving methods that could achieve the same task. 
A problem solving method decomposes a task into sub-tasks, and it might specify control information such as the order the sub-tasks should be executed in or whether they are required or not for achieving the goal of the parent task. 
The resulting structure is a directed, acyclic graph of tasks and problem solving methods and the leaf nodes of this graph are problem solving methods, called actions, that are explicit algorithms for how to solve a task.
Consider the generic task of designing an experiment. 
The input to the task would be the scientific theory and the hypothesis to be tested, and the output would be the research protocol containing the prediction, the study plan and the analysis plan in addition to the hypothesis.
How exactly to come up with the outputs would be specified in a problem solving method decomposing the parent task into the subtasks make study plan, make analysis plan and make prediction. 
Control information could be that the study plan would have to be made before the analysis plan, but that the prediction could be done in parallel with the other tasks as it only depends on the scientific theory. 
Although the subtasks and problem solving methods clearly specify how to complete the top-most task, the actions describes the exact steps that must be taken to perform a task. 
The actions are implemented in a programming language so that they can be executed on a computer running on some hardware. 

Figure~\ref{fig:hand-written_digit_calssification} illustrates a naive and simplistic task tree that shows the task-decomposition of the classic experiment of comparing hand-written digits using the MNIST data-set \cite{Lecun_1998}. 
Model comparison is a common empirical methodology in AI.
Different AI methods or machine learning models are implemented to solve the same task, such as classifying hand-written digits as the numbers they represent.  
When comparing hand-written digit classifiers, one need to collect data. 
This can be done by gathering some writers, make them write the numbers on paper and digitise the numbers by making one image for each digit.
Some pre-processing might be needed to improve performance.
The subtasks specified in the figure are the same pre-processing steps as used to make the MNIST data-set. 
After pre-processing, the images are ready to be classified. 
First, however, some classification algorithms might be selected, these should be the state of the art and maybe a new or a variation of an algorithm proposed by the investigators. 
The algorithms must be trained, hyperparameters must be optimized and they are tested on the test data.
In the MNIST experiment, the analysis is done by computing the test error although other performance metrics could be used, such as accuracy, F1-score and similar. 
The analysis is interpreted by comparing the test errors, and the lowest error indicate the best performing algorithm.
The interpretation of the result will be used to update our beliefs as illustrated in figure~\ref{fig:sientific_method-cs}.

A detailed research protocol could specify all tasks that are to be completed as part of an experiment and how to complete them by describing the more detailed steps, similar to a problem solving method. 
However, the amount of details that are needed to describe everything that is required to conduct a relatively complex experiment is huge, and this is not necessarily possible in writing. 
For example, according to Tian et al \cite{Tian_2019}, who tried to reproduce the results achieved by AlphaGo \cite{silver_2017} and Alpha Zero \cite{Silver_2018}, several details were missing from the first paper that hindered them in achieving their goal. 
Sometimes a paper contains pointers to possible implementations of how a sub-task of the experiment could be implemented, but not necessarily exactly how it was implemented. 
Differences in implementations could lead to different results. 
For example, Henderson et al. \cite{Henderson_2018} found that the result varied significantly for different implementations of \emph{the same} algorithm. 
As a task can be achieved by different problem solving methods, which problem solving method exactly is selected to complete a task, could affect the result.
For example, the problem solving method used for selecting which algorithms to compare could affect the result. 
Are the algorithms chosen based on whether they are the state of the art or because they are easiest for the researchers conducting the study to implement?
Also, the order that tasks are executed in might affect the result, so the control information of a problem solving method is important as well. 
It is not obvious that the result would be the same if the order of the pre-processing sub-tasks \emph{identify center of mass} and \emph{translate to center} are switched in the MNIST example.

\section{Reproducibility}
In this section, reproducibility is defined.
No distinction is made between reproducibility and replication, reflecting how Schmidt \cite{Schmidt_2009}, Nosek and Lakens \cite{Nosek_2014}, Goodman et al. \cite{Goodman_2016} and Gundersen and Kjensmo \cite{gundersen_kjensmo_2018} did not distinguish between the two concepts. 
Reproducibility is defined as follows: 
\begin{definition}
Reproducibility is the ability of independent investigators to draw the same conclusions from an experiment by following the documentation shared by the original investigators.
\end{definition}
Hence, reproducibility requires that another, independent team of investigators have to conduct the same experiment. 
This is contrasted with repeatability which is the ability of the same investigators to produce the same result when repeating an experiment. 
Also, an experiment is reproduced if the interpretation of the analysis leads to the same conclusions; the hypothesis that was supported by the original experiment is still supported after the independent investigators have conducted the same experiment. 
This means that results can be reproduced even when the outcome of the reproducibility experiment differ from the outcome of the original experiment, as long as the analysis can be interpreted in the same way and lead to the same conclusions.
Similarly, the analysis can also be different as long as the interpretation of the analysis leads to the same conclusion. 
From this follows that there are three degrees of reproducibility:
\begin{description}
    \item \textbf{Outcome reproducible:} 
    The outcome of the reproducibility experiment is the same as 
    the outcome produced by the original experiment. 
    When the outcome is the same, the same analysis and interpretation can be made, which leads to the same result and hence the hypothesis is supported by both experiments. 
    The experiment is outcome reproducible.
    \item \textbf{Analysis reproducible:}
    The outcome of the reproducibility experiment does not have to be the same as the outcome produced by the original experiment, but as long as the same analysis can be made and it leads to the same interpretation, the experiment is analysis reproducible. 
    \item \textbf{Interpretation reproducible:}
    Neither the outcome nor the analysis need to be the same as long as the interpretation of the analysis leads to the same conclusion. 
    In this case the experiment is interpretation reproducible.
\end{description}
If the outcome is not shared by the original investigators, independent researchers cannot compare their outcome with the outcome of the original experiment, and thus outcome reproducible is not achievable in such cases. 
In cases where researchers do not agree on the interpretation of an analysis, the results are not conclusive and the hypothesis is neither supported nor refuted. 
There is no new evidence to strengthen nor to weaken the belief that the hypothesis was formulated to test. 
New experiments must be conducted in order decide whether the hypothesis should be supported or refuted.

\begin{figure}[!t]
\centering\includegraphics[width=5in]{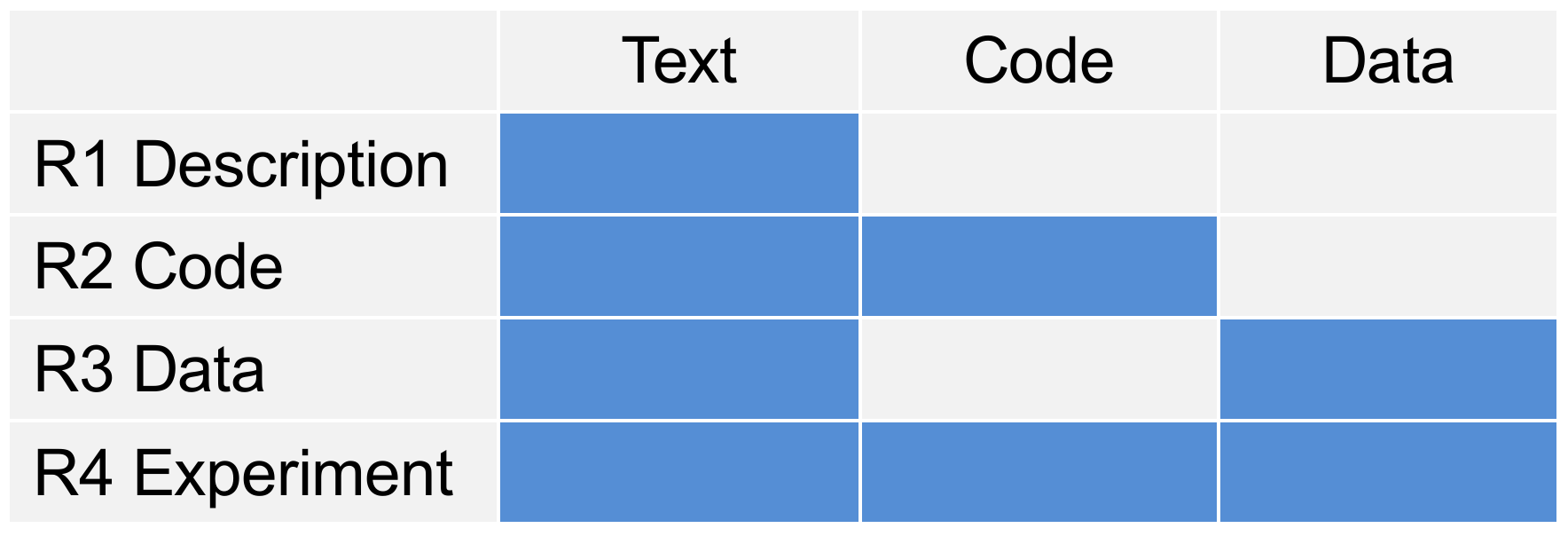}
\caption{The four reproducibility types R1 Description, R2 Code, R3 Data and R4 Experiment are defined by the types of documentation that is shared by the original investigators.}
\label{fig:transparency_degrees}
\end{figure}

A reproducibility experiment can only be conducted if some documentation of the original experiment is shared. 
The degree to which independent investigators can conduct the same experiment as the original investigators is dependent on how well the original investigators  documented the experiment and how much of the documentation is shared with independent investigators. 
Here, documentation is interpreted in a broad sense and is not restricted to textual descriptions similar to how Gundersen and Kjensmo \cite{gundersen_kjensmo_2018} interpreted documentation. 
They distinguish between three different types of documentation: text, code and data, and they propose to define reproducibility degrees based on which documentation types are shared with independent investigators. 
The term degree refers in this paper to the degree of closeness of the exact result that is reproduced while reproducibility type refers to which type of documentation is used for redoing the experiment.
The four reproducibility types are defined based on the documentation that is used to conduct the reproducibility experiment: 
\begin{description}
    \item \textbf{R1 Description:}
    Only a textual descriptions of the experiment is used as reference for the reproducibility experiment.
    The text could describe the experimental procedure, the target system and its behaviour, the implementation of the target system for example in form of pseudo code, the data collection procedure, the data, the outcome and the analysis and so on.
    \item \textbf{R2 Code:}
    Code and the textual description of the experiment are used as reference when conducting the reproducibility experiment.
    The code could cover the target system, the workflow, data pre-processing, experiment configurations, visualization and analyses. 
    \item \textbf{R3 Data:}
    Data and the textual description of the experiment are used as reference for conducting the experiment.
    The data could include training, validation and test sets as well as the outcome produced in the experiment.
    \item \textbf{R4 Experiment:}
    The complete documentation of the experiment including data and code in addition to the textual description as shared by the original investigators are used as reference for the reproducibility experiment.
\end{description}
Figure~\ref{fig:transparency_degrees} illustrates which documentation is required for the four reproducibility types. 
The more documentation that is shared means a higher degree of transparency and the easier it gets for independent investigators to reproduce the results.
However, the more variability in the conditions of the experiment, the more certain one can be that the conclusion is correct.
An experiment executed with the same code on the same data validates the hypothesis for specific code executed on specific data while a new implementation executed on different data generalizes the result to be independent of both code and data.
So, the less documentation that the original investigators share, the more increases the generality of the results. 
See Gundersen et al. \cite{gundersen_et_al_2018} for a discussion on the relationship between transparency and generality of results. 

Reproducibility types and degrees can be used to classify reproducibility experiments. 
A reproducibility experiment could be classified as \textit{OR4}, meaning that the outcomes were the same for  the original experiment and the reproducible experiment and that the reproducibility experiment was conducted by independent investigators based on the text, code and data shared by the original investigators.
\textit{IR1} would mean that the interpretation of the analysis of the reproducibility experiment lead to the same conclusion as the original experiment and that the reproducibility experiment was conducted by the independent investigators based on the textual description of the original experiment that was shared by the original investigators.

It is easier to reimplement a part of the experiment in order to introduce some difference, than having to implement the whole experiment.
Consider the task of comparing hand-written digit classifiers as illustrated in Figure~\ref{fig:hand-written_digit_calssification}.
For someone conducting this experiment, full transparency would be to share all the code and data that was used to conduct the experiment as well as the text describing the experiment. 
The research protocol could be considered the task and problem solving methods, while the implemented experiment is the actions. 
One could imagine that some researchers were not able to share their code, but they could share the task and problem solving methods, and hence the exact way they conducted the experiment, but not the code that implemented it.
This is basically what one seeks with the scientific article that describes the experiment. 
However, written language is not very precise and often there are limits to how long a scientific paper is allowed to be. 
Hence, written language is a poor replacement for code and higher level abstractions such as tasks and problem solving methods. 
If full transparency is provided, then one task, action or problem solving method can be reimplemented or changed to investigate whether the conclusion still is valid. 

Running the same experiment in the same environment is not considered to introduce any variability and is hence interpreted as repeatability.
Also, other investigators executing a program that encodes the same experiment on the same hardware means for computational sciences, as mentioned before, only that someone else pushes a button. 
This has little if any value.
Reproducing an experiment by executing the same code and data on a different computer with different ancillary software introduce variability though.

\section{Conclusion}
I started out by asking "what is reproducibility?", and I have spent the rest of the article investigating this question.
My strong belief is that this question could only be answered by understanding and describing how we discover knowledge by following the scientific method.
I have not seen anyone else approach the concept of reproducibility like this before.
This is quite surprising, as reproducibility is inherently related to experiments and the gold standard for conducting experiments is defined by the scientific method.
I have tried to analyse and characterise the scientific method to identify the most relevant concepts and used these concepts to analyse the literature on reproducibility. 

The literature on reproducibility do to a large degree agree that the same experiment is conducted as long as the same experimental method is followed.
However, following the same method is not enough.
An experiment is not reproduced unless the results are the same.
So what is the same result then?
The same result leads to the same conclusion. 
Hence, the hypothesis that was supported by the original experiment should be supported by the reproducibility experiment.
A cause of confusion, however, is the usage of the term result. 
It seems to mean both the outcome generated from  conducting the experiment as well as the interpretation of the analysis. 
I solved this by using the terms outcome, analysis and interpretation. 
The interpretation leads us to update our beliefs. 
Furthermore, reproducibility requires variation; a reproducibility experiment requires both sameness and difference. 
The sameness must at least be the same experimental methodology. 

The analysis resulted in a new definition of reproducibility and three degrees of reproducibility. 
Furthermore, four reproducibility types that are specified based on which types of documentation that are shared by the original investigators were proposed.
The more documentation that is shared, the easier it is for independent researchers to reproduce the results. 
The easier it is to reproduce results the faster knowledge can be discovered.
The understanding of reproducibility described here emphasizes that fast-paced and steady scientific progress require transparency and openness.













\bibliographystyle{RS}
\bibliography{bibliography}

\section{Appendix A: Supplementary Material}


\section*{Supplementary material}

| Paper title | Electronic documents give reproducible research a new meaning. |
|---|---|

| Reference | Claerbout 1992 |
|---|---|
| Term to be evaluated | reproducibility |

| Variable | Value | Evaluation |
|---|---|---|
| Investigators | 0 | Anyone with access to the CD ROM with the experiment can reproduce the work, so no distinction, but can be different. |
| Sci. theory | 2 | Supports the same theory as the code implements the same experiment. |
| Hypothesis | 2 | Evaluates the same theory as the code implements the same experiment. |
| Exp. procedure | 2 | The method is the same as it is implemented as code that is shared. |
| Implementation | 2 | The implementation is the same as the code is shared. |
| Ancillary | 0 | No restrictions on ancillary software. The code is not necessarily executed on the same software (version, OS). |
| Hardware | 0 | No restrictions to which hardware the code runs on, as anyone can execute on their own machine. |
| Data | 2 | The data is shared and used for the re-executing the experiment. |
| Outcome | 2 | The outcome is assumed to be the same. |
| Analysis | 2 | The analysis is the same, as visualizations and performance metrics that are calculated in code are the same. |
| Interpretation | 2 | The interpretation is assumed to be the same, as the figures are recomputed based on code and software, and the interpretation is typically stated in the text. ANALYSIS: In a case where the differences in computation affected by difference in software and hardware change the figures so much that they could be interpreted in a different way, the textual interpretation will be different from what the figures show. So, the interpretation if documented as text in the electronic document does not necessarily have to be the same as what should be the interpretation after looking at the recomputed figures. |

**Definition**
A revolution in education and technology transfer follows from the marriage of word processing and software command scripts. In this marriage an a.uthor attaches to every figure caption a push button or a name tag usable to recalculate the figure from all its data, parameters, and programs. This provides a concrete definition of reproducibility in computationally oriented research.

**Paper title**      WaveLab and Reproducible Research

**Reference**             Buckhet 1995
**Term to be evaluated**  reproducibility

| Variable | Value | Evaluation |
|---|---|---|
| Investigators | 0 | Anyone with access to shared WaveLab package containing the experiment can reproduce the work, so does not have to be, but can be different. |
| Sci. theory | 2 | Supports the same theory as the code implements the same experiment. |
| Hypothesis | 2 | Evaluates the same theory as the code implements the same experiment. |
| Exp. procedure | 2 | The method is the same as it is implemented as code that is shared. |
| Implementation | 2 | The implementation is the same as the code is shared. |
| Ancillary | 0 | Requires Matlab, but no restrictions on other ancillary software. The code is not necessarily executed on the same software (version, OS). |
| Hardware | 0 | No restrictions to which hardware the code runs on, as anyone can execute on their own machine. |
| Data | 2 | The data is shared as part of the WaveLab package and used for the re-executing the experiment. |
| Outcome | 2 | The outcome is assumed to be the same when running the experiment |
| Analysis | 2 | The analysis is the same, as visualizations and performance metrics that are calculated in code are the same. |
| Interpretation | 0 | The interpretation does not necessarily have to be the same. Interpretation is left for the investigator re-executing the experiment. |

**Definition**
Moreover, even those who make software available would rarely take the trouble to create a relatively comprehensive computing environment around it merely to ease the task of others wishing to reproduce the results. But that is exactly what we have done - constructed a whole computing environment aimed at allowing others to easily reproduce the figures in our articles. Our main goal in this paper is to call to attention to a principle we have tried to follow: When we publish articles containing figures which were generated by computer, we also publish the complete software environment which generates the figures. Comment: "The definition relies heavily on Claerbout 1992, which is reflected in the scores".

| | |
|---|---|
| **Paper title** | Reproducible Epidemiologic Research |
| | |
| **Reference** | Peng 2006 |
| **Term to be evaluated** | replication |

| Variable | Value | Evaluation |
|---|---|---|
| Investigators | 0 | From article: Scientific evidence is strengthened when findings are strengthened when replicated by multiple independent investigators |
| Sci. theory | 2 | Same experiment support same theory. |
| Hypothesis | 2 | Same findings support same hypothesis. |
| Exp. procedure | 0 | The method or study design, can be different |
| Implementation | 0 | The selection of subjects, handling of non-respondents and many other facets of the study can be different |
| Ancillary | 0 | Any software systems in different laboratories that the experiment relies on can be different. |
| Hardware | 0 | Laboratories and computer system hardware or other tools can be different. |
| Data | 0 | Different subjects are selected for the study. |
| Outcome | 0 | Outcome will be different. This is the data the article mentions as the outcome is what the experiment produces. |
| Analysis | 0 | Analytical methods can be different |
| Interpretation | 2 | The interpretation must be the same for the findings to be replicated. |

**Definition**

Scientific evidence is strengthened when important findings are replicated by multiple independent investigators using independent data, analytical methods, laboratories, and instruments. [...] Among the issues that cannot be addressed by reproducible research are those arising from stages of the research process prior to the data collection stage. Questions about the study design, the selection of subjects, the handling of nonrespondents, and many other facets of a study cannot be resolved with the analytical data alone. Similarly, it may not be possible to examine all relevant modeling choices, particularly those involving variables for which no data were originally collected. However, when we discuss these types of issues, we are moving closer to calling for full replication. If a particular study is fully replicable, then all aspects of the original study can be adjusted or modified. [Comment: The last sentence pinpoints the main aspect of the definition.]

**Paper title**    Reproducible Epidemiologic Research

**Reference**    Peng 2006
**Term to be evaluated**    reproducibility

| Variable | Value | Evaluation |
|---|---|---|
| Investigators | 0 | The analsis of the same data (outcome) is done by independent investigators. |
| Sci. theory | 2 | Investigating the same theory. |
| Hypothesis | 2 | The same hypothesis (finding) is tested. |
| Exp. procedure | 2 | As the same outcome is analysed by independent researchers the outcome is produced using the same method. |
| Implementation | 2 | As the same outcome is analysed by independent researchers the outcome is produced using the same implementation. |
| Ancillary | 2 | As the same outcome is analysed by independent researchers the outcome is produced using the same laboratory (and hence same ancillary software). |
| Hardware | 2 | As the same outcome is analysed by independent researchers the outcome is produced using the same laboratory (and hence same hardware). |
| Data | 2 | The same subjects are selected for the study. |
| Outcome | 2 | The analysis is done on the same data (outcome). |
| Analysis | 0 | A different analysis is done on the same data. |
| Interpretation | 2 | The intepretation of the different analysis must be the same for the independent investigators to replicate the finding. |

**Definition**
An attainable minimum standard is reproducibility, where independent investigators subject the original data to their own analyses and interpretations. Reproducibility calls for data sets and software to be made available for 1) verifying published findings, 2) conducting alternative analyses of the same data, 3) eliminating uninformed criticisms that do not stand up to existing data, and 4) expediting the interchange of ideas among investigators. Ultimately, all scientific evidence should be held to the standard of full replication and the confirmation of important findings by independent investigators. However, the desire to quantify small health effects and the significant weight placed on epidemiologic findings in the policy-making process create a need for epidemiologic studies to meet a minimum standard. We propose reproducibility to be this minimum standard.

**Paper title**      Replicability is not Reproducibility: Nor is it Good Science

**Reference**         Drummond 2009
**Term to be evaluated**   reproducibility

| Variable | Value | Evaluation |
|---|---|---|
| Investigators | 0 | Can be different, although researchers having problems with reproducing their own results is being mentioned. |
| Sci. theory | 2 | Supports the same theory, given example of speed of light. |
| Hypothesis | 2 | Supports the same hypothesis, given example of speed of light. |
| Exp. procedure | 0 | Experiment could be quite different, given example of speed of light |
| Implementation | 0 | As above. |
| Ancillary | 0 | As above. |
| Hardware | 0 | As above. |
| Data | 0 | As above. |
| Outcome | 0 | As above. |
| Analysis | 0 | As above. |
| Interpretation | 2 | Must be the same to support the same hypothesis. |

**Definition**

Reproducibility requires changes; replicability avoids them. [...] The lesson, I believe, that can be drawn from such an example is that Bradley obtained the result from quite a different experiment. In fact, apart from the result there seem few similarities. The original experiment certainly wasn't replicated. In fact, the claim that the speed of light was finite would not have been accepted by the community, in general, unless the experiment was very different. [...] In fact, I would claim that the greater the difference from the first experiment, the greater the power of the second.

**Paper title**          Replicability is not Reproducibility: Nor is it Good Science

**Reference**            Drummond 2009
**Term to be evaluated** replicability

| Variable | Value | Evaluation |
|---|---|---|
| Investigators | 0 | Can be different, although researchers having problems with reproducing their own results is being mentioned. |
| Sci. theory | 2 | As the code is shared, the experiment is the same and supports the same theory. |
| Hypothesis | 2 | As the code is shared, the experiment is the same and supports the same hypothesis. |
| Exp. procedure | 2 | As the code is shared, the experiment is the same experiment procedure is followed |
| Implementation | 2 | Code is shared. |
| Ancillary | 0 | Code can be executed on other software, no requirements to keep the same OS or software version. |
| Hardware | 0 | As above |
| Data | -1 | Sharing of data is not mentioned |
| Outcome | -1 | Will be different if data is different |
| Analysis | -1 | As long as done in code (metrics and visualization) it will be the same, but this is not specified. It is not clear whether the code that analyse the outcome should be shared as well, or only the code that implements the experiment. |
| Interpretation | 2 | Should be the same to support the hypothesis |

**Definition**
Reproducibility requires changes; replicability avoids them. [...] [Replicability] would be quite easy to achieve in machine learning simply by sharing the full code used for experiments. [...] It seems clear that authors believe that there should be no differences between one experiment and its reproduction. [Comment: The last sentence seems to indicate that the data should be shared as well, although this is not mentioned explicitly.]

| | | |
|---|---|---|
| **Paper title** | Shall We Really Do It Again? The Powerful Concept of Replication Is Neglected in the Social Sciences | |
| | | |
| **Reference** | Schmidt 2009 | |
| **Term to be evaluated** | direct replication | |

| Variable | Value | Evaluation |
|---|---|---|
| Investigators | 0 | This is not mentioned explicitly, but refers to Popper and that anyone should be able to test the observation (interpreted as redo the experiment). |
| Sci. theory | 2 | Narrow definition of repeating a procedure to support a hypothesis, which supports a theory. |
| Hypothesis | 2 | See scientific theory. |
| Exp. procedure | 2 | Definition is to repeat a procedure. The procedure is documented by a "recepie" so that anyone can follow it. |
| Implementation | -1 | Interpretation: The implementation (how exactly steps in the recepie are conducted) will most probaly differ, as it will be nearly impossible to document all parts of a protocol in social sciences as a recepie. However, this perspective (the difference between a procedure/recepie and its implementation) of an implementation is not covered and mentioned explicitly. |
| Ancillary | -1 | If someone else does it, the lab will be different and so the ancillary software, but not mentioned explicitly. Could easily be interpeted as 0. |
| Hardware | -1 | If someone else does it, the lab will be different and so the hardware but not mentioned explicitly. Could easily be interpeted as 0. |
| Data | 0 | In social sciences the participants differ as mentioned in definition (see below). |
| Outcome | 0 | When participants differ, the outcomes will differ, see the discussion on participants and how they will be different persons when time has passed, so redoing the same experiment with the same participants is the same as redoing the experiment with different participants. |
| Analysis | -1 | What exactly is meant by same results is not specified, neither is whether the analysis must be the same. |
| Interpretation | 2 | The interpretation must be the same to support the hypothesis. |

**Definition**
Therefore, I suggest differentiating at a fundamental level between two basic notions of replication: 1. Narrow bounded notion of replication - Repetition of an experimental procedure. Henceforth, this notion will be termed direct replication. [...] Whereas a direct replication is able to produce facts a conceptual replication may produce understanding. From fig 1: Scientific gain: fact confirmation and knowledge extension, Risk: Low risk. [Refers to social sciences and selection of participants:] As there are no two identical experiments this problem also applies to direct replications, although the possible sources of the failure to replicated a far more limited in this case. [Participants are interpreted as data and the output is the data that is produced by the experiment].

| Paper title | Shall We Really Do It Again? The Powerful Concept of Replication Is Neglected in the Social Sciences |
|---|---|

| Reference | Schmidt 2009 |
|---|---|
| Term to be evaluated | conceptual replication |

| Variable | Value | Evaluation |
|---|---|---|
| Investigators | 0 | This is not mentioned explicitly, but refers to Popper and that anyone should be able to test the observation (interpreted as redo the experiment). |
| Sci. theory | 2 | Repetiation of a test of a hypothesis that supports a theory, so the theory is the same. |
| Hypothesis | 2 | Same hypothesis is tested. |
| Exp. procedure | 0 | Different procedure. |
| Implementation | 0 | Hence difference implementation. |
| Ancillary | -1 | If someone else does it, the lab will be different and so the ancillary software, but not mentioned explicitly. Could easily be interpeted as 0. |
| Hardware | -1 | If someone else does it, the lab will be different and so the hardware but not mentioned explicitly. Could easily be interpeted as 0. |
| Data | 0 | In social sciences the participants differ as mentioned in definition (see below). |
| Outcome | 0 | When participants differ, the outcomes will differ, see the discussion on participants and how they will be different persons when time has passed, so redoing the same experiment with the same participants is the same as redoing the experiment with different participants. |
| Analysis | 0 | What exactly is meant by same results is not specified, neither is whether the analysis must be the same. |
| Interpretation | 2 | The interpretation must be the same to support the hypothesis |

**Definition**
Therefore, I suggest differentiating at a fundamental level between two basic notions of replication: 2. Wider notion of replication - Repetition of a test of a hypothesis or a result of earlier research work with different methods. Henceforth, this will be referred to as a conceptual replication. [...] Whereas a direct replication is able to produce facts a conceptual replication may produce understanding. [...] Thus, the function of conceptual replication is not only to confirm facts but also to assist in developing models and theories of the world. From fig 1. Scientific gain: understanding, Risk: High risk.

**Paper title**    Statistics and Chemometrics for Analytical Chemistry.

**Reference**    Miller 2010
**Term to be evaluated**    repeatability

| Variable | Value | Evaluation |
|---|---|---|
| Investigators | 2 | Same students doing same experiment over again. |
| Sci. theory | 2 | Same students doing same experiment over again. |
| Hypothesis | 2 | Same students doing same experiment over again. |
| Exp. procedure | 2 | Same students doing same experiment over again. |
| Implementation | 2 | Same students doing same experiment over again. |
| Ancillary | 2 | In same lab using same tools (SW and HW). |
| Hardware | 2 | In same lab using same tools (SW and HW). |
| Data | 2 | Same solutions or phenomena being studies. |
| Outcome | 2 | The outcome would be different as there is a within run precision. |
| Analysis | 2 | Same analysis, done for all the runs. |
| Interpretation | 2 | As it is the same experiment repeated to generate new outcomes, the interpretation would be made based on these outcomes and would hence be the same for all of them. The procedure is repeated to make the outcome that the analysis and interpreation is based on. |

**Definition**
Another important area of terminology is the difference between reproducibility and repeatability. We can illustrate this using the students' results again. In the normal way each student would do the five replicate titrations in rapid succession, taking only an hour or so. The same set of solutions and the same glassware would be used throughout, the same preparation of indicator would be added to each titration flask, and the temperature, humidity and other laboratory conditions would remain much the same. In such cases the precision for each student would be the within-run precision: this is called the repeatability.

| | | |
|---|---|---|
| **Paper title** | Statistics and Chemometrics for Analytical Chemistry. | |
| | | |
| **Reference** | Miller 2010 | |
| **Term to be evaluated** | reproducibility | |

| Variable | Value | Evaluation |
|---|---|---|
| Investigators | -1 | What is important in this definition is not that someone else does the experiment, but that is done in a different lab (it could be by someone else or the same investigators). So the physical lab is what is relevant, not the people conducting the experiment. |
| Sci. theory | 2 | Same experiment supports the same theory and hypothesis. |
| Hypothesis | 2 | Same experiment supports the same theory and hypothesis. |
| Exp. procedure | 2 | Same experiment means it should follow the same procedure. |
| Implementation | -1 | The implementation (how exactly steps in the experimental procedure are conducted) will most probaly differ, as it will be nearly impossible to document all parts of a protocol in chemistry - workflow ie.. However, this perspective (the difference between a procedure/recepie and its implementation) of an implementation is not covered and mentioned explicitly |
| Ancillary | 0 | Different laboratories means different tools and probably different software. |
| Hardware | 0 | Different laboratories means different tools and different hardware. |
| Data | 0 | Different laboratories means different tools and probably materials or solutions. |
| Outcome | 0 | When using different solutions, materials in a different lab the outcome will be different, indicated by the term between run precision. |
| Analysis | -1 | Whether the results (outcome) should be analysed in the same way is not specified. |
| Interpretation | -1 | To support the same hypthesis the same interpretation must be made, but this is not mentioned explicitly. Someone could get very different outcomes and therefore the between run precision shows that they hypothesis is wrong. |

**Definition**
Suppose, however, that for some reason the titrations were performed by each student on five different occasions in different laboratories, using different pieces of glassware and different batches of indicator. It would not be surprising to find a greater spread of the results in this case. The resulting data would reflect the between-run precision of the method, i.e. its reproducibility. [...] The reproducibility of a method is normally expected to be poorer (i.e. with larger random errors) than its repeatability.

**Paper title**      Reproducible Research in Computational Science

**Reference**           Peng 2011
**Term to be evaluated**   publication only

| Variable | Value | Evaluation |
|---|---|---|
| Investigators | 0 | Independence is mentioned in the ingress. |
| Sci. theory | 2 | Not mentioned, but implicit as hypothesis related to scientific theory. |
| Hypothesis | 2 | Hypothesis mentioned in relation to replication. |
| Exp. procedure | 2 | This is what the publication describes. |
| Implementation | 0 | Not shared when publication only. |
| Ancillary | 0 | Not shared when publication only. |
| Hardware | 0 | Will differ when publication only. |
| Data | 0 | Will differ when publication only. |
| Outcome | 0 | Will differ when publication only. |
| Analysis | -1 | Might be the same, as long as analysis described in the publication, but this is not mentioned. |
| Interpretation | 2 | Interpretation must be the same to support the hypothesis. |

**Definition**
Reproducibility in contrast to full replication: "Reproducibility has the potential to serve as a minimum standard for judging scientific claims when full independent replication of a study is not possible." [Never stated explicitly, but assumed that "Not reproducible" in spectrum requires full replication]. Not reproducible is when "publication only" is available [see spectrum Fig 1].

| | | |
|---|---|---|
| **Paper title** | Reproducible Research in Computational Science | |

| | | |
|---|---|---|
| **Reference** | Peng 2011 | |
| **Term to be evaluated** | full replication | |

| Variable | Value | Evaluation |
|---|---|---|
| Investigators | 0 | Independence is mentioned in the ingress. |
| Sci. theory | 2 | Same experiment supports the same theory and hypothesis. |
| Hypothesis | 2 | Same experiment supports the same theory and hypothesis. |
| Exp. procedure | 2 | This is what the publication describes. |
| Implementation | 2 | Code is shared so implementation is the same. |
| Ancillary | 0 | Might be different. |
| Hardware | 0 | Might be different. |
| Data | 2 | Shared so it is the same. |
| Outcome | 2 | The associate editor for reproducibility runs the submitted code on the data and verifies that the code produces the results published in the article. What exactly is meant by results is not unambiguous. |
| Analysis | 2 | Might be the same, as long as analysis described in the publication, but this is not mentioned. |
| Interpretation | 2 | Interpretation must be the same to support the hypothesis. |

**Definition**

Authors can submit their code or data to the journal for posting as supporting online material and can additionally request a "reproducibility review," in which the associate editor for reproducibility runs the submitted code on the data and verifies that the code produces the results published in the article.

**Paper title**      Trust Your Science? Open Your Data and Code

**Reference**      Stodden 2011
**Term to be evaluated**      replicability

| Variable | Value | Evaluation |
|---|---|---|
| Investigators | 0 | Regenerate from author-provided code, so indicates other investigator than the author. |
| Sci. theory | 2 | Same experiment supports the same theory and hypothesis. |
| Hypothesis | 2 | Same experiment supports the same hypothesis |
| Exp. procedure | 2 | Defined by shared code. |
| Implementation | 2 | Code is shared. |
| Ancillary | 0 | Will differ when experiment conducted by someone else. |
| Hardware | 0 | Will differ when experiment conducted by someone else. |
| Data | 2 | Data is shared. |
| Outcome | 2 | Regeneration of published results from author-provided code and data. |
| Analysis | 2 | As above. |
| Interpretation | 2 | Has to be the same to support the hypothesis. |

**Definition**

I learned from my advisor, David Donoho, that reproducibility meant releasing data and code such that others may regenerate your results on their own systems (i.e., releasing the full computational environment that produces a result). In our case, this typically meant releasing MATLAB scripts and data files, often along with a MATLAB GUI we wrote that permitted the user to select the figure he or she wished to regenerate, adjust parameter settings, and view the source code (www-stat.stanford.edu/~wavelab, http://sparselab.stanford.edu). Unless we can inspect the code and data, we cannot resolve differences in output between independent methods or independent implementations of even purportedly identical methods. We can reserve the term "replicability" for the regeneration of published results from author-provided code and data. [...] [T]hat sufficient information exists with which to understand, evaluate, and build upon a prior work if a third party could replicate the results without any additional information from the author.

| Paper title | Trust Your Science? Open Your Data and Code |
|---|---|

| Reference | Stodden 2011 |
|---|---|
| **Term to be evaluated** | reproducibility |

| Variable | Value | Evaluation |
|---|---|---|
| Investigators | 0 | Other investigator than the author. |
| Sci. theory | 2 | Same experiment supports the same theory and hypothesis. |
| Hypothesis | 2 | Same experiment supports the same theory and hypothesis. |
| Exp. procedure | 2 | Same experiment, but with some independence. |
| Implementation | 0 | Not using same code. |
| Ancillary | 0 | Will differ when experiment conducted by someone else. |
| Hardware | 0 | Will differ when experiment conducted by someone else. |
| Data | 0 | Not using same data. |
| Outcome | 0 | Differs when using different code and data. |
| Analysis | 0 | Might differ. |
| Interpretation | 2 | Has to be the same to support the hypothesis and regenerate the findings. |

**Definition**

Reproducibility is a more general term, implying both replication and the regeneration of findings with at least some independence from the code and/or data associated with the original publication.

| Paper title | International vocabulary of metrology – Basic and general concepts and associated terms |
|---|---|

| Reference | JCGM 2012 |
|---|---|
| **Term to be evaluated** | repeatability |

| Variable | Value | Evaluation |
|---|---|---|
| Investigators | 2 | Same opertors doing same experiment over again. |
| Sci. theory | -1 | Not specified. |
| Hypothesis | -1 | Not specified. |
| Exp. procedure | 2 | Same measurement procedure. |
| Implementation | -1 | Not specified. |
| Ancillary | 2 | Same operating conditions. |
| Hardware | 2 | Same measuring system, interpreted as hardware. |
| Data | 1 | Same or similar objects. |
| Outcome | 0 | The outcome would be different as there is a within run precision. |
| Analysis | -1 | Analysis is not covered by this term. |
| Interpretation | -1 | Interpretation is not covered by this term. The term does not say something about the result or drawing the same conclusions from it. |

**Definition**

Repeatability condition of measurement repeatability condition: Condition of measurement, out of a set of conditions that includes the same measurement procedure, same operators, same measuring system, same operating conditions and same location, and replicate measurements on the same or similar objects over a short period of time [Interpreted as keeping everything the same].

**Paper title**     International vocabulary of metrology – Basic and general concepts and associated terms

**Reference**     JCGM 2012
**Term to be evaluated**     reproducibility

| Variable | Value | Evaluation |
|---|---|---|
| Investigators | 0 | Out of a set of conditions that includes different operators. |
| Sci. theory | -1 | Not specified. |
| Hypothesis | -1 | Not specified. |
| Exp. procedure | 2 | Same measurements procedure. |
| Implementation | -1 | Not specified. |
| Ancillary | 0 | Out of a set of conditions that includes different measuring systems. |
| Hardware | 0 | Out of a set of conditions that includes measuring systems. |
| Data | 1 | Out of a set of conditions that includesreplicate measurements on the same or similar objects. |
| Outcome | 0 | Measurements are expected to be different with some variation. |
| Analysis | -1 | Analysis is not covered by this term. |
| Interpretation | -1 | Interpretation is not covered by this term. The term does not say something about the result or drawing the same conclusions from it. |

**Definition**
Reproducibility condition of measurement reproducibility condition: Condition of measurement, out of a set of conditions that includes different locations, operators, measuring systems, and replicate measurements on the same or similar objects. [..] A specification should give the conditions changed and unchanged, to the extent practical.

**Paper title**     Learning from the Past: Approaches for Reproducibility in Computational Neuroscience

**Reference**     Crook 2013
**Term to be evaluated**     internal replicability

| Variable | Value | Evaluation |
|---|---|---|
| Investigators | 2 | The quest for reproducibility raises the question of what it means to reproduce a result independently. |
| Sci. theory | 2 | Same experiment supports the same theory and hypothesis. |
| Hypothesis | 2 | Same experiment supports the same hypothesis. |
| Exp. procedure | 2 | Defined by shared code. |
| Implementation | 2 | Code is shared. |
| Ancillary | 2 | Same lab, same SW (implied by stating different HW/SW in external). |
| Hardware | 2 | Same lab, same HW (see above). |
| Data | -1 | Simulation, so no data. |
| Outcome | 0 | Independent reproduction of experimental results will necessarily entail deviations from the original experiment, as not all conditions can be fully controlled. Whether a result is considered to have been reproduced successfully thus requires careful scientific judgement, differentiating between core claims and mere detail in the original report. |
| Analysis | 2 | Outcome might differ, but everything else same, also analysis. |
| Interpretation | 2 | Has to be the same to support the hypothesis. |

**Definition**
The original author or someone in their group can re-create the results in a publication, essentially by rerunning the simulation software. For complete replicability within a group by someone other than the original author, especially if simulations are performed months or years later, the author must use proper bookkeeping of simulation details using version control and electronic lab journals. [...] There are some intermediate steps between pure internal replication of a result (the original author rerunning the original code)

**Paper title**    Learning from the Past: Approaches for Reproducibility in Computational Neuroscience

**Reference**    Crook 2013
**Term to be evaluated**    external replicability

| Variable | Value | Evaluation |
|---|---|---|
| Investigators | 0 | The quest for reproducibility raises the question of what it means to reproduce a result independently. |
| Sci. theory | 2 | Same experiment supports the same theory and hypothesis. |
| Hypothesis | 2 | Same experiment supports the same hypothesis. |
| Exp. procedure | 2 | Defined by shared code. |
| Implementation | 2 | Code is shared. |
| Ancillary | 0 | External replicability may be sensitive to the use of different operating systems, compilers, and libraries. |
| Hardware | 0 | External replicability may be sensitive to the use of different hardware. |
| Data | -1 | Simulation, no data. |
| Outcome | 0 | Independent reproduction of experimental results will necessarily entail deviations from the original experiment, as not all conditions can be fully controlled. Whether a result is considered to have been reproduced successfully thus requires careful scientific judgement, differentiating between core claims and mere detail in the original report. |
| Analysis | 2 | Same code. |
| Interpretation | 2 | Has to be the same to support the hypothesis. |

**Definition**
A reader is able to re-create the results of a publication using the same tools as the original author. As with internal replicability, all implicit knowledge about the simulation details must be entered into a permanent record and shared by the author. This approach also relies on code sharing, and readers should be aware that external replicability may be sensitive to the use of different hardware, operating systems, compilers, and libraries. [...] and fully independent reproduction, namely external replication (someone else rerunning the original code).

**Paper title**  Learning from the Past: Approaches for Reproducibility in Computational Neuroscience

**Reference**  Crook 2013
**Term to be evaluated** cross-replicability

| Variable | Value | Evaluation |
|---|---|---|
| Investigators | 0 | The quest for reproducibility raises the question of what it means to reproduce a result independently. |
| Sci. theory | 2 | Same experiment supports the same theory and hypothesis. |
| Hypothesis | 2 | Same experiment supports the same hypothesis. |
| Exp. procedure | 2 | Defined by shared code. |
| Implementation | 1 | Access to code while reimplementing or running it on different simulation platform capable of running same simulation independent code. |
| Ancillary | 0 | As for external replicability: may be sensitive to the use of different operating systems, compilers, and libraries. |
| Hardware | 0 | As for external replicability: may be sensitive to the use of different hardware |
| Data | -1 | Simulation, no data. |
| Outcome | 0 | Independent reproduction of experimental results will necessarily entail deviations from the original experiment, as not all conditions can be fully controlled. Whether a result is considered to have been reproduced successfully thus requires careful scientific judgement, differentiating between core claims and mere detail in the original report. |
| Analysis | -1 | Questions about how to compare results. Not explict. |
| Interpretation | 2 | Has to be the same to support the hypothesis. |

**Definition**
The use of "cross" here refers to simulating the same model with different software. This may be achieved by re-implementing a model using a different simulation platform or programming language based on the original code, or by executing a model described in a simulator-independent format on different simulation platforms. Simulator-independent formats can be divided into declarative and procedural approaches. Assuming that all simulators are free of bugs, this would expose the dependence of simulation results on simulator details, but leads to questions about how to compare results. [...] cross-replication (running cross-platform code on a different simulation platform).

**Paper title**    Learning from the Past: Approaches for Reproducibility in Computational Neuroscience

**Reference**    Crook 2013
**Term to be evaluated**    reproducibility

| Variable | Value | Evaluation |
|---|---|---|
| Investigators | 0 | The quest for reproducibility raises the question of what it means to reproduce a result independently. |
| Sci. theory | 2 | Same experiment supports the same theory and hypothesis. |
| Hypothesis | 2 | Same experiment supports the same hypothesis. |
| Exp. procedure | 2 | From reading paper. |
| Implementation | 0 | Independent implementation, so different, not looking at code even. |
| Ancillary | 0 | Different. |
| Hardware | 0 | Different. |
| Data | -1 | Simulation, no data. |
| Outcome | 0 | Will differ. |
| Analysis | -1 | Not specified. |
| Interpretation | 2 | Has to be the same to support the hypothesis. |

**Definition**

Bob reads Alice's paper, takes note of all model properties, and then implements the model himself using a simulator of his choice. He does not download Alice's scripts. Bob's implementation thus constitutes an independent implementation of the scientific ideas in Alice's paper based on a textual description. The boundary line between cross-replicability and reproducibility is not always clear. In particular, a declarative description of a model is a structured, formalized version of what should appear in a publication, so an implementation by Charlie based on a declarative format might be considered to be just as independent as that by Bob based on reading the article. [...] re-implementing a model with knowledge of the original code.

**Paper title**    Recomputation.org: Experiences of Its First Year and Lessons Learned

**Reference**    Gent 2014
**Term to be evaluated**    recomputation

| Variable | Value | Evaluation |
|---|---|---|
| Investigators | 0 | The site is intended to serve as a repository for computational experiments, embodied in virtual machines so that they can be recomputed at will by other researchers. |
| Sci. theory | 2 | Same experiment supports the same theory and hypothesis. |
| Hypothesis | 2 | Same experiment supports the same hypothesis. |
| Exp. procedure | 2 | Defined by shared code. |
| Implementation | 2 | Code is shared. |
| Ancillary | 2 | Same as in virtual machine. |
| Hardware | 0 | VM can be run on different hardware. |
| Data | 2 | Same as stored in virtual machine. |
| Outcome | 0 | Recomputation is not equal to Exact Reproduction of Results. Instead, we argue that in reality we must expect to get different results every time we run an experiment, even when we recompute using the identical machine each time. |
| Analysis | -1 | If in code, then same, but not specified, as recompuation is not about getting the same result over and over again. This is mentioned several times. |
| Interpretation | -1 | Recomputation is about enabling executing the code again, not to support hypotheses. |

**Definition**

By recomputation we mean the act of embodying a computational experiment in a virtual machine so that it can be preserved and rerun at any point in the future. [...] We do not see recomputation as the same thing as exact reproduction of results. Indeed, this goes beyond varying measures such as runtime, and can affect results obtained such as the predicted weather in one day's time.

**Paper title**    Registered Reports - A Method to Increase the Credibility ofPublished Results

**Reference**    Nosek 2014
**Term to be evaluated**    direct replication

| Variable | Value | Evaluation |
|---|---|---|
| Investigators | 0 | This differentiates science from other ways of knowing – confidence in claims is not based on trusting the source, but in evaluating the evidence itself. |
| Sci. theory | 2 | Experiment supports same theory. |
| Hypothesis | 2 | Same experiment tests the same hypothesis |
| Exp. procedure | 2 | A direct replication is the attempt to duplicate the conditions and procedure that existing theory and evidence anticipate as necessary for obtaining the effect. |
| Implementation | 1 | A direct replication is the attempt to duplicate the conditions existing theory and evidence anticipate as necessary for obtaining the effect. |
| Ancillary | 1 | A direct replication is the attempt to duplicate the conditions existing theory and evidence anticipate as necessary for obtaining the effect. |
| Hardware | 1 | A direct replication is the attempt to duplicate the conditions existing theory and evidence anticipate as necessary for obtaining the effect. |
| Data | 0 | Direct replications add data (so it is different). |
| Outcome | 0 | Different data leads to different outcome. |
| Analysis | -1 | Not mentioned. |
| Interpretation | 2 | Supports same hypothesis. |

**Definition**
First, direct replications add data to increase precision of the effect size estimate via meta-analysis. Under some circumstances, this can lead to the identification of false positive research findings. Without direct replication, there is no way to confidently identify false positives. Second, direct replication can establish generalizability of effects. There is no such thing as an exact replication. Any replication will differ in innumerable ways from the original. Third, direct replications that produce negative results facilitate the identification of boundary conditions for real effects. If existing theory anticipates the same result should occur and, with a high-powered test, it does not, then something in the presumed irrelevant differences between original and replication could be the basis for identifying constraints on the effect. In other words, understanding any effect requires knowing when it does and does not occur. Therefore, replications and negative results are consequential for theory development. A direct replication is the attempt to duplicate the conditions and procedure that existing theory and evidence anticipate as necessary for obtaining the effect [...] In principle, the evidence supporting scientific knowledge can be reproduced by following the original methodologies. This differentiates science from other ways of knowing – confidence in claims is not based on trusting the source, but in evaluating the evidence itself.

| Paper title | Registered Reports - A Method to Increase the Credibility of Published Results |
| --- | --- |

| Reference | Nosek 2014 |
| --- | --- |
| Term to be evaluated | conceptual replication |

| Variable | Value | Evaluation |
| --- | --- | --- |
| Investigators | 0 | This differentiates science from other ways of knowing – confidence in claims is not based on trusting the source, but in evaluating the evidence itself. |
| Sci. theory | 2 | Experiment supports same theory. |
| Hypothesis | 2 | Same experiment tests the same hypothesis. |
| Exp. procedure | 0 | Different procedure. |
| Implementation | 0 | Different procedure means different implementation. |
| Ancillary | 0 | Abstract a phenomenon from its original operationalization. |
| Hardware | 0 | Abstract a phenomenon from its original operationalization. |
| Data | 0 | Abstract a phenomenon from its original operationalization. |
| Outcome | 0 | Different data leads to different outcome. |
| Analysis | -1 | Not mentioned. |
| Interpretation | 2 | Must be same to support same hypothesis. |

**Definition**

Conceptual replications have been more popular than direct replications because they abstract a phenomenon from its original operationalization and contribute to our theoretical understanding of an effect. However, conceptual replication are not best suited to clarify the truth of any particular effect because nonsignificant findings are attributable to changes in the research design, and rarely lead researchers to question the phenomenon.

**Paper title**          What does research reproducibility mean?

**Reference**            Goodman 2016
**Term to be evaluated** results reproducible

| Variable | Value | Evaluation |
| --- | --- | --- |
| Investigators | 0 | The ability of a researcher to duplicate the results of a prior study. |
| Sci. theory | 2 | Experiment supports same theory. |
| Hypothesis | 2 | Same experiment tests the same hypothesis. |
| Exp. procedure | 2 | Ability to implement, as exactly as possible, the experimental procedurs. |
| Implementation | 2 | Ability to implement, as exactly as possible, the computational procedures. |
| Ancillary | 1 | Ability to implement, as exactly as possible, the experimental procedurs. |
| Hardware | 1 | With the same  tools. |
| Data | 2 | With the same data. |
| Outcome | 2 | To obtain the same results. |
| Analysis | 2 | To obtain the same results. |
| Interpretation | 2 | To obtain the same results. |

**Definition**
Methods reproducibility is meant to capture the original meaning of reproducibility, that is, the ability to implement, as
exactly as possible, the experimental and computational procedures, with the same data and tools, to obtain the same results.

**Paper title**          What does research reproducibility mean?

**Reference**            Goodman 2016
**Term to be evaluated** methods reprodcuible

| Variable | Value | Evaluation |
|---|---|---|
| Investigators | 0 | The ability of a researcher to duplicate the results of a prior study. |
| Sci. theory | 2 | Experiment supports same theory. |
| Hypothesis | 2 | Same experiment tests the same hypothesis. |
| Exp. procedure | 2 | Following the same experimental methods. |
| Implementation | 0 | Different. |
| Ancillary | 0 | Different. |
| Hardware | 0 | Different. |
| Data | 0 | Different. |
| Outcome | 0 | Different. |
| Analysis | 0 | Different. |
| Interpretation | 2 | Must support the same hypothesis to corroborate the results. |

**Definition**
Results reproducibility refers to what was previously described as "replication," that is, the production of corroborating results in a new study, having followed the same experimental methods.

**Paper title**    What does research reproducibility mean?

**Reference**    Goodman 2016
**Term to be evaluated**    inferential reproducible

| Variable | Value | Evaluation |
|---|---|---|
| Investigators | 0 | The ability of a researcher to duplicate the results of a prior study. |
| Sci. theory | 2 | Experiment supports same theory. |
| Hypothesis | 2 | Same experiment tests the same hypothesis. |
| Exp. procedure | 2 | Ability to implement, as exactly as possible, the experimental procedures. |
| Implementation | 2 | Ability to implement, as exactly as possible, the computational procedures. |
| Ancillary | 1 | With the same  tools. |
| Hardware | 1 | With the same  tools. |
| Data | 2 | SWith the same data and tools. |
| Outcome | 2 | To obtain the same results. |
| Analysis | 0 | This is not identical to results reproducibility, because not all investigators will draw the same conclusions from the same results, or they might make different analytical choices that lead to different inferences from the same data. |
| Interpretation | 2 | Will be the same for the experiment to be inferentially reproducible. |

**Definition**
Inferential reproducibility, not often recognized as a separate concept, is the making of knowledge claims of similar strength from a study replication or reanalysis. This is not identical to results reproducibility, because not all investigators will draw the same conclusions from the same results, or they might make different analytical choices that lead to different inferences from the same data [Same as results reproducibilty, except for analysis, which might be different. This means that if interpretation is different between original and reproduced experiments, the experiment is still reproduced, but with different conclusion.]

**Paper title**      Artifact Review and Badging – Version 1.0

**Reference**      ACM 2020
**Term to be evaluated**      repeatability

| Variable | Value | Evaluation |
|---|---|---|
| Investigators | 2 | For computational experiments, this means that a researcher can reliably repeat her own computation. |
| Sci. theory | 2 | Experiment supports same theory. |
| Hypothesis | 2 | Same experiment tests the same hypothesis. |
| Exp. procedure | 2 | Same procedure as same code. |
| Implementation | 2 | Same code. |
| Ancillary | 2 | On same software system. |
| Hardware | 2 | On same hardware |
| Data | 2 | Using same data. |
| Outcome | 2 | Reliably repeat her own computation. |
| Analysis | 2 | Same analysis. |
| Interpretation | 2 | Same interpretation. |

**Definition**
Repeatability (Same team, same experimental setup): The measurement can be obtained with stated precision by the same team using the same measurement procedure, the same measuring system, under the same operating conditions, in the same location on multiple trials. For computational experiments, this means that a researcher can reliably repeat her own computation. [...] For example, artifacts can be software systems, scripts used to run experiments, input datasets, raw data collected in the experiment, or scripts used to analyze results.

| Paper title | Artifact Review and Badging – Version 1.0 |
| --- | --- |

| Reference | ACM 2020 |
| --- | --- |
| **Term to be evaluated** | reproducibility |

| Variable | Value | Evaluation |
| --- | --- | --- |
| Investigators | 0 | independent group. |
| Sci. theory | 2 | Experiment supports same theory. |
| Hypothesis | 2 | Same experiment tests the same hypothesis. |
| Exp. procedure | 2 | Same procedure as same code (Artifact). |
| Implementation | 2 | Using the author's own artifacts. |
| Ancillary | 0 | Different ancillary software. |
| Hardware | 0 | Different hardware. |
| Data | 2 | Using the author's own artifacts. |
| Outcome | 0 | Will be different, although similar. |
| Analysis | 2 | If done in code: using the author's own artifacts. |
| Interpretation | 2 | Obtain the same result. |

**Definition**

Reproducibility (Different team, different experimental setup ): The measurement can be obtained with stated precision by a different team using the same measurement procedure, the same measuring system, under the same operating conditions, in the same or a different location on multiple trials. For computational experiments, this means that an independent group can obtain the same result using the author's own artifacts. [...] For example, artifacts can be software systems, scripts used to run experiments, input datasets, raw data collected in the experiment, or scripts used to analyze results.

| Paper title | Artifact Review and Badging – Version 1.0 |
|---|---|

| Reference | ACM 2020 |
|---|---|
| **Term to be evaluated** | replicability |

| Variable | Value | Evaluation |
|---|---|---|
| Investigators | 0 | Different team. |
| Sci. theory | 2 | Experiment supports same theory. |
| Hypothesis | 2 | Same experiment tests the same hypothesis. |
| Exp. procedure | 2 | Same procedure from documentation that is shared by original team. |
| Implementation | 0 | Using artifacts which they develop completely independently. |
| Ancillary | 0 | Different ancillary software. |
| Hardware | 0 | Different hardware. |
| Data | 0 | Using artifacts which they develop completely independently. |
| Outcome | -1 | Probably not what is meant by results, will with high probability be different. |
| Analysis | -1 | Same analysis? If in code it would be developed indpendently, but not mentioned. |
| Interpretation | 2 | Same result. |

**Definition**

Replicability (Different team, same experimental setup ): The measurement can be obtained with stated precision by a different team, a different measuring system, in a different location on multiple trials. For computational experiments, this means that an independent group can obtain the same result using artifacts which they develop completely independently.

**Paper title**          State of the Art: Reproducibilty in Artificial Intelligence

**Reference**            Gundersen 2018
**Term to be evaluated** experiment reproducible

| Variable | Value | Evaluation |
|---|---|---|
| Investigators | 0 | Independent research team. |
| Sci. theory | 2 | Same experiment supports the same theory and hypothesis. |
| Hypothesis | 2 | Same experiment supports the same hypothesis. |
| Exp. procedure | 2 | Defined by shared code. |
| Implementation | 2 | Code is shared. |
| Ancillary | 0 | Different. |
| Hardware | 0 | Different. |
| Data | 2 | Data is shared. |
| Outcome | -1 | Not specified. |
| Analysis | -1 | Not specified. |
| Interpretation | 2 | Same interpretation, see fig 1. |

**Definition**

Reproducibility in empirical AI research is the ability of an independent research team to produce the same results using the same AI method based on the documentation made by the original research team. [...] The results of an experiment are experiment reproducible when the execution of the same implementation of an AI method produces the same results when executed on the same data.

| | | |
|---|---|---|
| **Paper title** | State of the Art: Reproducibilty in Artificial Intelligence | |

| | | |
|---|---|---|
| **Reference** | Gundersen 2018 | |
| **Term to be evaluated** | data reproducible | |

| Variable | Value | Evaluation |
|---|---|---|
| Investigators | 0 | Independent research team. |
| Sci. theory | 2 | Same experiment supports the same theory and hypothesis. |
| Hypothesis | 2 | Same experiment supports the same hypothesis. |
| Exp. procedure | 2 | Textual description of experiment shared. |
| Implementation | 0 | Code is not shared. |
| Ancillary | 0 | Different. |
| Hardware | 0 | Different. |
| Data | 2 | Data is shared. |
| Outcome | -1 | Not specified. |
| Analysis | -1 | Not specified. |
| Interpretation | 2 | Same interpretation, see fig 1. |

**Definition**

Reproducibility in empirical AI research is the ability of an independent research team to produce the same results using the same AI method based on the documentation made by the original research team. [...] R2: Data Reproducible - The results of an experiment are data reproducible when an experiment is conducted that executes an alternative implementation of the AI method that produces the same results when executed on the same data.

**Paper title**      State of the Art: Reproducibilty in Artificial Intelligence

**Reference**      Gundersen 2018
**Term to be evaluated**      method reproducible

| Variable | Value | Evaluation |
| --- | --- | --- |
| Investigators | 0 | Independent research team. |
| Sci. theory | 2 | Same experiment supports the same theory and hypothesis. |
| Hypothesis | 2 | Same experiment supports the same hypothesis. |
| Exp. procedure | 2 | Textual description of experiment shared. |
| Implementation | 0 | Code is not shared. |
| Ancillary | 0 | Different. |
| Hardware | 0 | Different. |
| Data | 0 | Data is not shared. |
| Outcome | -1 | Not specified. |
| Analysis | -1 | Not specified. |
| Interpretation | 2 | Same interpretation, see fig 1. |

**Definition**

Reproducibility in empirical AI research is the ability of an independent research team to produce the same results using the same AI method based on the documentation made by the original research team. [...] R3: Method Reproducible - The results of an experiment are method reproducible when the execution of an alternative implementation of the AI method produces the same results when executed on different data.

| | |
|---|---|
| **Paper title** | Reproducibility and Replicability in Science (2019) |
| | |
| **Reference** | NAS 2019 |
| **Term to be evaluated** | reproducibility |

| Variable | Value | Evaluation |
|---|---|---|
| Investigators | 0 | … a second researcher follows the methods described by the first researcher … |
| Sci. theory | 2 | Same hypothesis supports same theory. |
| Hypothesis | 2 | Same experiment supports same experiment. |
| Exp. procedure | 2 | When results are produced by complex computational processes using large volumes of data, the methods section of a scientific paper is insufficient to convey the necessary information for others to reproduce the results. |
| Implementation | 2 | Obtaining consistent computational results using the same computational steps, methods, and code. |
| Ancillary | 1 | The computational environment should be documented so that independent researchers can use similar (from recommendation 4-1). |
| Hardware | 1 | As above. |
| Data | 2 | Obtaining consistent computational results using the same input data. |
| Outcome | 2 | Obtaining consistent computational results. |
| Analysis | 2 | Obtaining consistent computational results using the same input data analysis. |
| Interpretation | 2 | Not explicitly stated, but implicitly if same result (outcome) is produced and same analysis made, then the interpretation will be the same and the results resproduced. |

**Definition**
We define reproducibility to mean computational reproducibility—obtaining consistent computational results using the same input data, computational steps, methods, and code, and conditions of analysis. […] In short, reproducibility involves the original data and code.

| | |
|---|---|
| **Paper title** | Reproducibility and Replicability in Science (2019) |
| | |
| **Reference** | NAS 2019 |
| **Term to be evaluated** | replicability |

| Variable | Value | Evaluation |
|---|---|---|
| Investigators | 0 | ... a second researcher follows the methods described by the first researcher ... |
| Sci. theory | 2 | Answering the same scientic question (hypothesis) would indicate same theory. |
| Hypothesis | 2 | Studies aimed at answering the same scientific question. |
| Exp. procedure | 0 | Different studies answering the same scientific questions. |
| Implementation | 0 | Different study. |
| Ancillary | 0 | Different study. |
| Hardware | 0 | Different study. |
| Data | 0 | Different study. |
| Outcome | 0 | Different study. |
| Analysis | 0 | Different study. |
| Interpretation | 2 | Obtaining consistent results across studies aimed at answering the same scientific question. |

**Definition**

We define replicability to mean obtaining consistent results across studies aimed at answering the same scientific question, each of which has obtained its own data. [...] In short, replicability involves new data collection and similar methods used by previous studies.



\end{document}